\documentclass[11pt]{elsarticle}

\usepackage[utf8]{inputenc}
\usepackage{amsmath,amssymb}
\usepackage{graphicx}
\usepackage[table,xcdraw]{xcolor}
\usepackage{svg}
\usepackage{tabularx}
\usepackage{multirow}
\usepackage{comment}
\usepackage{algorithm}
\usepackage{algpseudocode}
\usepackage{subfig}
\definecolor{codegreen}{rgb}{0,0.6,0}
\definecolor{codegray}{rgb}{0.5,0.5,0.5}
\definecolor{codepurple}{rgb}{0.58,0,0.82}
\definecolor{backcolour}{rgb}{0.95,0.95,0.92}

\usepackage{dsfont}
\usepackage{mathtools}
\usepackage{pdflscape}
\usepackage{nicefrac,geometry}
\usepackage{hyperref}

\begin{document}

\begin{frontmatter}

\title{A Benchmarking Suite for Flexible Job Shop Scheduling Problems with Worker Flexibility under Uncertainty}

\author[inst1]{David Hutter}

\affiliation[inst1]{organization={Research Centre Business Informatics, Josef Ressel Centre for Robust Decision Making,\\ Vorarlberg University of Applied Sciences},
            addressline={Hochschulstrasse 1},
            city={Dornbirn},
            postcode={6850},
            country={Austria}
            }

\author[inst1]{Thomas Steinberger}
\author[inst1]{Michael Hellwig}

\begin{abstract}
This paper addresses the Flexible Job Shop Scheduling Problem and its extension with Worker Flexibility, which integrates workforce assignment into machine-operation scheduling. Diverse solvers have been proposed across multiple optimization domains including Mathematical Programming, Constraint Programming, and Simulation-Based Optimization, or Simulation-based Optimization. These are often tailored to narrow use cases and validated on limited test problem sets, hindering cross-domain comparison. To overcome this, a comprehensive benchmarking environment built on 402 standardized Flexible Job Shop Scheduling Problem instances is introduced and systematically extended to include worker flexibility. This creates a hitherto unique collection of ready-to-use worker flexibility instances. The benchmark suite features several metrics for algorithm performance assessment, the visualization of algorithmic results, as well as state-of-the-art baseline results. This enables rigorous, reproducible, and comparable performance analysis between solvers and scheduling problem subdomains. Through the simulation-based integration of uncertainties in processing times as well as resource availabilities, the environment supports the development and evaluation of robust optimization strategies. The present work lays a foundation for targeted algorithm development and consistent performance evaluation in production scheduling research.
\end{abstract}

\begin{highlights}
\item Design of a predefined benchmark collection for FJSSP with Worker Flexibility.
\item Optional inclusion of different kinds of scheduling uncertainties.
\item Presentation of an open-source benchmarking suite for scheduling solver assessment.
\item Accessibility and comparability for solvers from multiple optimization paradigms.
\item Support for algorithm development by instance filtering and result harmonization.

 \end{highlights}


\end{frontmatter}

\section{Introduction}
\label{sec:introduction}

\begin{figure}[b]\centering
\includegraphics[width=0.9\textwidth]{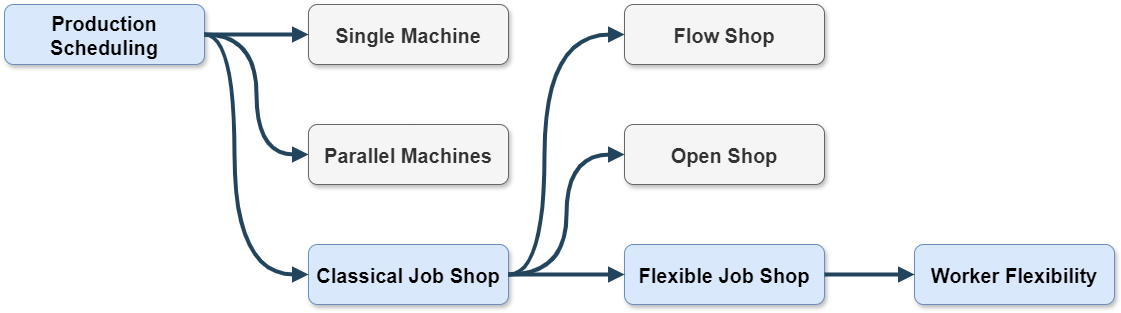}
    \caption{Visualization of problem classes in Production Scheduling. The blue boxes indicate the branch considered in this paper. Notice that other problem classes exist and this illustration is no means  exhaustive.}
    \label{fig:schedulingTax}
\end{figure}

Planning problems occur across many economic domains where efficient resource allocation is essential. Robust solutions are vital in sectors such as Supply Chain Management, Energy, Healthcare, and Finance, influencing sustainable growth, operational efficiency, and overall competitiveness~\cite{xie2019review,GHASEMI2024100599,dauzere-peres_flexible_2024}.
In Manufacturing, Production Scheduling (PS) allocates resources such as machinery, labor, and materials to meet production targets efficiently. This includes meeting delivery times, minimizing idle times, and responding to machine breakdowns. PS problems involve the determination of a proper sequence of  decisions to attain (near) optimal performance on the shop floor. They take a variety of forms and are commonly approached by applying techniques from Artificial Intelligence (AI), Operations Research (OR), or Simulation-based Optimization (SBO)~\cite{GHASEMI2024100599}.  
A broad overview of common PS problem classes is provided in Fig.~\ref{fig:schedulingTax}. The focus of the present work is on the Flexible Job-Shop Scheduling Problem (FJSSP) and its extension to worker flexibility (FJSSP-W).

In PS, the diversity of methodological approaches and problem representations makes it difficult to assess the quality of algorithmic performance and calls for a consistent comparison of state-of-the-art optimizers~\cite{ghasemi_simulation_2024}. A corresponding benchmarking framework is crucial for advancing research and practice in the field of simulation-based optimization for production scheduling~\cite{bartzbeielstein2020benchmarkingoptimizationbestpractice}. As the optimization problems become increasingly complex - incorporating machine flexibility, workforce constraints, and sources of uncertainty - solvers from different optimization domains each may offer complementary strengths~\cite{whitley1996evaluating,PIOTROWSKI2023101378}. 
However, without a standardized benchmarking framework, comparisons of their effectiveness remain fragmented and anecdotal~\cite{HELLWIG2019bench,PIOTROWSKI2025101807}. A rigorous benchmarking environment enables fair, reproducible evaluation across different problem instances and solution approaches, revealing trade-offs in performance, scalability limits, and robustness under different conditions~\cite{rardin2001experimental,McGeoch2008,johnson2002experimental}. This not only serves as a guide for algorithm development and selection, but also promotes cross-domain innovation and helps researchers and practitioners identify the most suitable approaches for real-world planning challenges~\cite{kononova2025benchmarkingmattersrethinkingbenchmarking}.
As far as the systematic benchmarking of algorithmic solvers for PS problems is concerned, only a few partial aspects are covered, particularly in the area of SBO methods. The development and performance comparison of such solvers particularly relies on benchmarking due to the lack of theoretical performance results for optimization tasks of notable complexity~\cite{bartzbeielstein2020benchmarkingoptimizationbestpractice}. First and foremost, benchmarking experiments are established for performance evaluation and algorithm comparison on given problem classes. Ideally, this is supposed to support the selection of the algorithm best suitable for given real-world applications~\cite{10.1162/EVCO_a_00134}. Further, benchmarks can be used to deliver insights into the working principles of algorithms and foster their development for specific problem types~\cite{more2009benchmarking,NIKOLIKJ2025101895}.

Benchmarking environments are recently developed for many problem classes, e.g. satisfiability problems~\cite{HONJOIDE2025101933}, path planning problems~\cite{DARLAN2025101968}, Traveling Sales Person problems~\cite{Weise2014}, general continuous single-objective constrained optimization~\cite{HELLWIG2019bench}, or dynamic optimization problems~\cite{LIN2022101184, FOX2022101125}, to name just a few. 
Yet, there are only few developments in the field of production scheduling. 
While~\cite{reijnen2025jobshopschedulingbenchmark} proposes a benchmark suite for Job Shop Scheduling Problems (JSSP), to the best of our knowledge, no systematic benchmarking environment exists for the FJSSP—especially none that incorporates worker flexibility or uncertainty factors. For the FJSSP the landscape remains fragmented into the following cases:

\emph{Sets of problem instances:} Over the years, multiple authors proposed their own versions of FJSSP test problem instance sets~\cite{behnke_test_2012, reijnen2025jobshopschedulingbenchmark} and new problem instances are constantly being created~\cite{march2024novel}.
Although, widely accepted and comprehensive collections of benchmark sets are openly available~\cite{noauthor_schedulinglabfjsp-instances_2025,noauthor_scheduling_benchmark_environments_and_instances_2025, leo_lei-kunfjsp-benchmarks_2025, behnke_test_2012}, the rather arbitrary use of instance subsets is the usual practice. 
This practice hinders solver comparability across studies and limits a comprehensive understanding of algorithmic working principles.

\emph{Real-world instances:} Many PS solvers are developed for specific real-world problems. While there are efforts to derive reasonable benchmarking environments from such specific industrial use cases~\cite{trentesaux_benchmarking_2013, ghasemi_simulation_2024}, the majority of cases does not allow for the (systematically) comparison of different solvers. This is partly due to the very individual requirements of these real-world problems that complicate the application of different solvers~\cite{peng_petri_2019}. On the other hand, the computational effort can be too extensive for a rigorous assessment, or details of the real-world problem dependencies might not be properly disclosed~\cite{ghasemi_simulation_2024}.

\emph{Competitions:} Conferences like GECCO and CEC 
regularly organize PS competitions~\cite{gecco2025_energy_competition,cec2025_dffsp_competition}, providing well‑documented and structured benchmarks. However, these competitions are often tailored to specific use cases with unique requirements. As a result, broader comparisons with solvers that are not adapted to these particularities become difficult.

\emph{Performance metrics:} A frequently discussed benchmarking aspect in PS is the definition of performance metrics~\cite{cavalieri_benchmarking_2007, turkyilmaz_research_2020}. These can vary greatly depending on the scheduling objective, application area and other influencing factors (noise, dynamic constraints, etc.)~\cite{hazir_robust_2010}. While a universal guideline is unrealistic, it remains essential to apply performance measures consistently when comparing algorithms.

When it comes to considering worker-related constraints, i.e. the FJSSP-W, the benchmarking landscape becomes even more scattered. There is neither a commonly accepted set of test instances nor a standardized method for creating such instances. Instead, researchers often develop test instances independently~\cite{ gnanavelbabu_simulation-based_2021,luo_distributed_2022} or rely on solitary problem representations from the real world~\cite{gong2020hybrid,gong2020energy}.
This lack of standardization undermines core benchmarking principles such as comparability and reproducibility, as documentation is often incomplete or key real‑world problem details remain undisclosed. It also hinders the integration of uncertainties into problem instances. Systematically incorporating stochastic processing times, machine failures, or random job arrivals~\cite{dauzere-peres_flexible_2024} requires a reliable set of noise‑free base instances to assess relative performance. Moreover, the corresponding uncertainty models must be designed in a replicable manner.

\textbf{Contribution:} 
Therefore, this work presents a closed, systematic benchmarking environment that addresses several of the shortcomings outlines above. 
It combines a comprehensive collection of FJSSP problem instances~\footnote{Those FJSSP instances from~\cite{dauzere-peres_flexible_2024} and~\cite{behnke_test_2012} are selected that meet the FJSSP definition and must not be assigned to more specialized problem classes (e.g. parallel machines, or flow store scheduling), cf. Fig.~\ref{fig:schedulingTax}.}~\cite{dauzere-peres_flexible_2024,behnke_test_2012} with a structured apporach for extending them to flexible workers settings~\cite{nouri_benchmark_2018}. In this way, we create a fixed portfolio of FJSSP-W instances. By specifying a consistent objective and providing an assessment methodology with coherent visualizations, the benchmarking suite enables a systematic, repeatable, and comparable solver evaluation.
Traceable modifications from FJSSP to FJSSP-W instances make it possible to understand the rise in complexity between problem classes and its impact on algorithm performance. Standardized test criteria and termination conditions additionally support a consistent benchmarking. 
Through optional user-specific filtering based on instance properties, the suite supports the analysis and development of algorithmic operators for specific problem groups (e.g., particularly high flexibility in the production line).
Selecting representative instances for particular characteristic also allows for repeated testing with reduced computational effort, without requiring evaluation on the full instance set.
The additional extension for uncertainty simulations enables the development and comparison of robust optimization approaches for creating resilient production schedules under uncertain conditions. Furthermore, it allows for the evaluation of the effects of uncertainties on the solver’s performance. To this end, the benchmarking environment offers the option of incorporating three sources of uncertainty into the problem instances: uncertainty in processing times, uncertainty in machine failures, and uncertainty in worker availability. The corresponding uncertainties are modeled based on realistic PS assumptions and integrated into the benchmarking problems via simulations.

Although designed primarily for the comparison and development of SBO algorithms, e.g. Evolutionary Algorithms (EA), the platform allows for the relative performance assessments of state-of-the-art solvers across optimization paradigms (like classical Mathematical Programming) by also providing the respective model representations for MILP and CP solvers within the documentation of the benchmarking repository~\cite{Hutter_Benchmarking}. Such representations also allow for reproduction of the best-known results available, which were obtained by applying multiple state-of-the-art MILP and CP solvers like Gurobi~\cite{gurobi}, Hexaly~\cite{noauthor_hexaly_nodate}, IBM CPLEX~\cite{noauthor_constraint_2024}, or OR-Tools~\cite{noauthor_or-tools_nodate}. Thus the benchmarking suite will be useful for comparison and selection of appropriate solvers for the specific problem classes. 

The remainder of this paper is organized as follows: Section~\ref{sec:relwork} provides the FJSSP problem formulation and an overview of available benchmark sets. The extension towards worker flexibility, i.e. FJSSP-W, is then specified in Sec.\ref{sec:workflex}. In Sec.\ref{sec:benchmark}, the structure of the benchmarking environment is described. This is followed by the demonstration of test solvers in Sec.\ref{sec:experiments}.
Finally, the paper concludes in Sec.\ref{sec:concl}, and proposes some directions for future work in Sec.\ref{sec:outlook}, respectively.

\section{The Flexible Job Shop Scheduling Problem}

\label{sec:relwork}
The FJSSP forms the foundation of the proposed benchmarking suite. It extends the classical Job‑Shop Scheduling Problem (JSSP), in which multiple jobs share resources but follow different processing routes. In contrast to JSSP, the FJSSP allows each operation to be processed on one of several eligible machines, each with potentially different processing times. This added flexibility reflects more dynamic PS environments. From a solver’s perspective, it increases problem complexity: algorithms must determine not only the optimal operation sequence but also the most suitable machine assignment for each operation.

Next, a formal definition of the FJSSP problem is presented in Sec.~\ref{sec:problem} together with helpful problem features that allow for a characterization of individual problem instances. Sec.~\ref{sec:benchmarks} then provides a brief overview of the available FJSSP benchmark instances.

\subsection{FJSSP Problem Formulation}
\label{sec:problem}

The FJSSP class consists of NP-complete combinatorial optimization problems that require to sequence a series of operations on a subset of available machines while minimizing a certain cost function $C$~\cite{10.5555/574848}.
The machines are not necessarily homogeneous, meaning that the processing time for each operation depends on the machine it is assigned to. 
Using the notation from~\cite{hutter_interior-point_2024}, the general FJSSP representation introduces a set of $n$ jobs $\mathcal{J}=\{J_1,\dots,J_n\}$ that need to be processed on a set of $m$ machines $\mathcal{M}=\{M_1,\dots,M_m\}$. 
Each of these jobs $J_i \: (i=1,\dots,n)$ consists of a predetermined sequence of $n_i$ operations. The total number of operations is thus calculated as $ \texttt{N} = \sum_{i=1}^n n_i$. Accordingly, the ratio of the total amount of operations and the number of jobs gives the average number of operations per job $\frac{\texttt{N}}{n}$. 
Moreover, the $j$th operation within the $i$th job $J_i$ is refered to as $O_{i,j}$. Each operations $O_{i,j}$ is performed by one of the $m_{i,j}$ admissible machines in the subset $\mathcal{M}_{i,j} \subset \mathcal{M}$. 
The constant processing time of operation $O_{i,j}$ on machine $M_k \in \mathcal{M}_{i,j} \: (k=1 ,\dots, m_{i,j})$ is given as $T_{i,j,k}$. The minimum processing duration among all possible operation-machine combinations of a single problem instance is referred to as ${T}_{\min}$. Conversely, the maximum processing duration is ${T}_{\max}$. Their difference defines the so-called duration span $\Delta T = {T}_{\max} - {T}_{\min}$. 
The general FJSSP representation assumes that all jobs are known and all machines are operational at time $0$. 

Diverse objective functions are conceivable for the FJSSP, the optimization of which affects the final production plan.
These include minimizing  queue times, maximizing throughput, minimizing tardiness, or maximizing resource utilization, to name but a few~\cite{dauzere-peres_flexible_2024}.
One of the most important objective functions across literature is the makespan, i.e. the total time required for the execution of all jobs of a problem instance. Hence, the makespan will be used as the default objective of the proposed FJSSP benchmarking environment in this work. However, it is intended to represent the objective function as general as possible (both in the problem modeling and in the performance analysis) to allow for the potential inclusion of other optimization objectives. 

Since every maximization task can be formulated equivalently as a minimization problem, w.l.o.g. it can be assumed that a given cost function is to be minimized in full satisfaction of the production constraints of the FJSSP. Consequently, the cost function represents a mapping of a certain 
 schedule representation $\mathbf{y} \in \mathcal{Y}$ to a particular cost function value $C(\mathbf{y})\in\mathds{R}$, i.e. $ C : \mathcal{Y} \to \mathds{R} $.
Hence, one is interested in the search space vector $\mathbf{y} $ that realizes a minimal $C(\mathbf{y})$ value. Note, that concrete examples of a parameter vector encoding for $\mathbf{y}$ is depending on the modeling approach and will be provided in the supplementary material of this work~\cite{Hutter_Benchmarking}.


\subsection{FJSSP benchmark instances}
\label{sec:benchmarks}

Although several FJSSP test collections exist, they are rarely used systematically when evaluating new solvers. Consequently, only a limited number of algorithms are benchmarked on broad sets of FJSSP instances~\cite{dauzere-peres_flexible_2024}. Most approaches are tested only on small subsets, introducing latent selection bias and limiting the comparability of results. Many studies also focus solely on individual real‑world cases~\cite{GHASEMI2024100599}, and the often highly specialized algorithms presented are seldom compared rigorously against established solvers on diverse instances. As a result, the strengths of genuinely strong methods may remain unnoticed, while minor improvements risk being overstated.

A comprehensive collection commonly used FJSSP instances is provided in~\cite{behnke_test_2012}. The paper presents a number of long-established test instances, proposes some self-developed ones, and reports on the corresponding best-known results. A more recent survey~\cite{dauzere-peres_flexible_2024} discusses common problem instances, and updates the best-known results. However, the investigated problem instances in both works do not overlap perfectly.



The proposed FJSSP benchmarking suite considers in total $402$ distinct FJSSP instances from ten test problem collections that are listed in~\cite{behnke_test_2012}, and~\cite{dauzere-peres_flexible_2024}, respectively. Notice that only those instances are selected that meet the FJSSP definition provided in Sec.~\ref{sec:problem}. Closely related instance collections in~\cite{birgin_milp_2014} and~\cite{demirkol_benchmarks_1998} are excluded for the following reasons: While the collection in~\cite{demirkol_benchmarks_1998} is limited to JSSP instances, i.e. it renounces the flexibility to run any of the operations on any machine, the instances in~\cite{birgin_milp_2014} allow parallel processing of some operations within a job which calls for a different model representation and exceeds the scope of the present FJSSP benchmark suite.

To ensure a better understanding of the differences between the individual instances and facilitate a systematic selection of user-preferred subgroups, characteristic properties for all 402 FJSSP instances are provided in the public Github repository of the benchmarking environment~\cite{Hutter_Benchmarking}. Along the total number of jobs ($n$), machines ($m$), or operations ($\texttt{N}$) of each instance, these characteristics include the minimum ($T_{\min}$), maximum ($T_{\max}$) and average processing times ($\bar{T}$) measured over all possible machine-operation combinations, as well as the corresponding standard deviation $\sigma_{T}$. Moreover, the average number of operations per job $\nicefrac{\texttt{N}}{n}$, the machine flexibility $\beta$, and the duration variety $dv$ are provided. The instance characteristics quantify the degree of flexibility and differences in processing times that an instance is subject to, thus reflecting an intuition for the complexity of the FJSSP benchmark instances~\cite{hutter_interior-point_2024}.
An overview of the FJSSP source collections, their number of instances, as well as some of their average characteristics is provided in Table~\ref{tab:benchmarksources}.
\begin{table}[t]
\centering
\caption{Overview of FJSSP benchmark problem collections from literature. The table provides the number of instances per source collection (\# inst) as well as their average characteristics. These include the number of jobs $\bar{n}$, the amount of operations $\bar{\texttt{N}}$, the average amount of operations per job $\bar{\texttt{N}}/\bar{n}$, the amount of machines $\bar{m}$, the flexibility $\bar{\beta}$ (cf. Eq.~\ref{beta_flexibility}), and the duration variety $\bar{dv}$ (see Eq.~\ref{duration_variety}). }
\label{tab:benchmarksources}
\scalebox{0.945}{
\renewcommand{\arraystretch}{1.1}
\begin{tabular}{|rl|r|r|r|r|r|r|r|}\hline
\textbf{No.} & \textbf{Source name} & \textbf{\# inst} & $\bar{n}$ & $\bar{\texttt{N}}$ & \textbf{$\nicefrac{\bar{\texttt{N}}}{\bar{n}}$} & $\bar{m}$ & $\bar{\beta}$ & $\bar{dv}$  \\\hline
1                                    & BehnkeGeiger \cite{behnke_test_2012}         & 60 & 45.00 & 225.00 & 5.0 & 40.0 & 0.316 & 0.009         \\
2                                    & Brandimarte  \cite{brandimarte_routing_1993}         & 15 & 20.33 &  171.87 & 8.56 & 9.13 & 0.310 & 0.007           \\
3                                    & ChambersBarnes \cite{barnes_solving_1995}       & 21 & 13.33 & 158.33 & 11.67 & 13.667 & 0.089 & 0.007            \\
4                                    & DPpaulli \cite{dauzere-peres_integrated_1997}              & 18 & 15.00    & 292.00 & 19.49 & 7.667 & 0.330 & 0.004                  \\
5                                    & Fattahi     \cite{fattahi_mathematical_2007}          & 20 & 5.35 & 17.40 & 2.95 & 5.1 & 0.517 & 0.098       \\
6                                    & Kacem  \cite{kacem_pareto-optimality_2002}               & 4 & 9.75 & 31.75 & 3.16 & 8.0 & 1.0 & 0.042                    \\
7                                    & Hurink EData    \cite{hurink_tabu_1994}      & 66 & 14.76 & 133.38 & 8.85 & 8.848 & 0.151 & 0.010                    \\
8                                    & Hurink SData  \cite{hurink_tabu_1994}        & 66 & 14.76 & 133.38 & 8.85 & 8.848 & 0.131 & 0.010      \\
9                                    & Hurink RData  \cite{hurink_tabu_1994}        & 66 & 14.76 & 133.38 & 8.85 & 8.848 & 0.258 & 0.010      \\
10                                   & Hurink VData  \cite{hurink_tabu_1994}        & 66 & 14.76 & 133.38 & 8.85 & 8.848 & 0.476 & 0.010    \\ \hline
\textbf{}   & \textbf{Overall} & \textbf{402} & \textbf{18.89} & \textbf{150.11} & \textbf{8.54} & \textbf{13.510} & \textbf{0.281} & \textbf{0.014}  \\\hline
\end{tabular}
}
\end{table}
In this context, the flexibility $\beta$ of an instance is defined as the ratio of the average amount of available machines per operation $m_{avg}$ and the total number of machines $m$, 
\begin{equation}
    \label{beta_flexibility}
    \begin{split}
        \beta = \cfrac{m_{avg}}{m}.
    \end{split}
\end{equation}
It indicates how many of the machines in the production environment can handle multiple operations~\cite{escamilla-serna_global-local_2021}.
In case, all operations of all jobs can be processed by any machine, $\beta =1$ and the FJSSP is called fully flexible. Otherwise, i.e. $0 < \beta < 1$,  it is only partially-flexible and, potentially, harder to solve.
The duration variety $dv$ is defined as the quotient of the number of unique values of the processing times $T_{i,j,k}$ within a FJSSP instance $d_{unique}$ and the total number of machine assignment options across all operations,
\begin{equation}
    \label{duration_variety}
    \begin{split}
        dv = \frac{d_{unique}}{\sum_{i,j}m_{i,j}}.
    \end{split}
\end{equation}
 Low $dv$ values indicate either very similar processing times across all available machines or a large amount of assignment options, respectively. This corresponds to many distinct scheduling solutions that have the same solution quality, i.e. equal makespan values.

\subsection{FJSSP instance description}
\begin{figure}[t]
   \centering
   \includegraphics[width=0.95\textwidth]{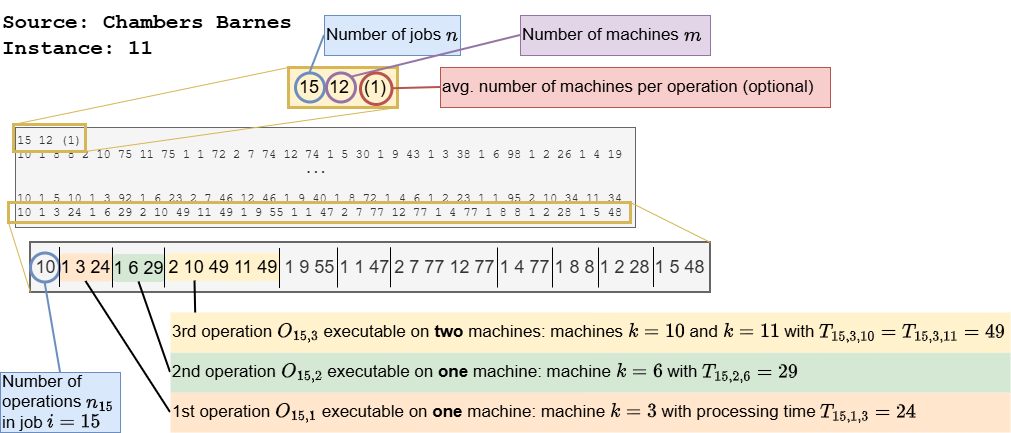}
   \caption{Explanation of the FJSSP specifications on the basis of the 11th test instance in~\cite{barnes_solving_1995}. }
   \label{fig:textfile}
\end{figure}

The usual specification of the production environments is given for each individual FJSSP instance in the form of a text file containing all necessary information about the dependencies in the manufacturing environment. An example is given in Fig.~\ref{fig:textfile} using the 11th instance of the FJSSP collection from~\cite{barnes_solving_1995} to make it easier for the less experienced reader to understand the specifications.

The first line of the text file contains two (optionally three) numbers: The first value specifies the number of jobs $n$ and the second value indicates the number of machines $m$ in the FJSSP instance. The optional third value gives the average number of machines per operation $m_{avg}$, but is not necessarily required for the implementation of the problem instance. In Fig.~\ref{fig:textfile}, the instance comes with $n=15$ jobs, $m=12$ machines available, and an average number of one machine per operation. The remaining rows then represent the individual jobs $J_i$ $(i=1,...,n)$ of the instance, each beginning with the number of operations $n_i$ to be executed in the job. 
The second number gives the number of machines $m_{i,1}$ available to process the first operation $O_{i,1}$. This is followed by a sequence of $m_{i,1}$ pairs of numbers $(k,T_{i,1,k})$ indicating the index $k$ of the admissible machines $M_k$ as well as the processing times $T_{i,1,k}$ of operation $O_{i,1}$ on machine $M_k$. Next, the data for the second operation $O_{i,2}$ are provided. In this way, all operations of all jobs are specified accordingly. Refer to Fig.~\ref{fig:textfile} for the detailed description of the first operations in job $J_{15}$.


\section{Incorporating Worker Flexibility and Uncertainty}
\label{sec:workflex}

In order to significantly increase operational efficiency in production environments, it is important to take employee flexibility into account in planning so that different skill and availability levels can be incorporated. 
This section looks at the integration of employee flexibility into the FJSSP. It provides a suitable problem formulation of the resulting Flexible Job Shop Scheduling Problem with Worker Flexibility (FJSSP-W), discusses the current benchmarking situation, and talks about the design of the FJSSP-W instance set. The optional inclusion of uncertainties in processing times by the workforce also enables measurement of uncertainty effects on the production process and performance assessment of robust planning procedures.

\subsection{FJSSP-W problem formulation and instance characteristics}

As an extension of the FJSSP, the FJSSP-W requires that an eligible worker is assigned to operate the machine to which an operation is assigned. The selected machine-worker combination further determines the processing time of the operation on the machine. Following the notations of Sec.~\ref{sec:problem} and~\cite{hutter_workerflex_2025}, the formulation of the FJSSP-W additionally introduces a set of $w$ workers $\mathcal{W}=\{W_1,\dots,W_{w}\}$. 
Consequently, each operation $O_{i,j}$ needs to be assigned to a permissible machine $M_{k} \in \mathcal{M}_{i,j} \subset \mathcal{M}$ and a worker $W_{l} \in \mathcal{W}_{i,j,k}$, where $\mathcal{W}_{i,j,k} \subset \mathcal{W}$ denotes the subset of $w_{i,j,k} = \left| \mathcal{W}_{i,j,k} \right|$ available workers to process the operation $O_{i,j}$  on the assigned machine $M_{k}$. The corrsponding processing time of operation $O_{i,j}$ on machine $M_k \in \mathcal{M}_{i,j}$ with worker $W_l \in W_{i,j,k}$ is given as $T_{i,j,k,l}$. Similar to the FJSSP, the FJSSP-W assumes that all operations, machines, and workers are operational and available at time $0$. The objective for the FJSSP-W is again to find those operation-machine-worker combinations that minimize the makespan $C$ of all jobs, i.e. the cumulative processing times of all operations. 

In contrast to the FJSSP, the introduction of worker dependencies demands for modifications of some of the instance characteristics introduced before. Particularly, the flexibility $\beta$ and duration variety $dv$, i.e., Eqs.~\eqref{beta_flexibility} and ~\eqref{duration_variety}, need adjustments to account for the additional worker requirements.
The flexibility $\beta$ of an FJSSP-W instance requires the consideration of all possible worker assignments for each operation-machine combination. It is given as the ratio of the average amount of machine-worker combinations per operation
\begin{equation}
    \label{eq:average_assignments}
    \omega_{avg} = \frac{1}{\texttt{N}} \sum_{i=1}^{n}\sum_{j=1}^{n_i}\sum_{k=1}^{m_{i,j}} w_{i,j,k},
\end{equation}
and the overall amount of unique machine-worker combinations available for the benchmark instance under consideration $\omega_{unique}$. Accordingly, the flexibility $\beta$ is obtained as
\begin{equation}
     \label{eq:beta_fjssp-w}
    \beta = \cfrac{\omega_{avg}}{\omega_{unique}}.
\end{equation}
The $\beta$ value of an FJSSP instance indicates how many workers can handle several machines. If all existing assignment options are available for all operations, one receives $\beta = 1$ and the FJSSP-W is fully flexible. Otherwise, one obtains $0<\beta<1$, which reflects only partial flexibility of an instance and is generally more difficult to solve.
The modified duration variety $dv$ is defined as the quotient of the number of unique processing times $T_{i,j,k,l}$ values of the FJSSP-W instance $d_{unique}$, and the total number of operation-machine-worker combinations across all operations $\omega_{avg} \cdot \texttt{N}$, i.e.,
\begin{equation}
    \label{eq:dv_fjssp-w}
        dv = \cfrac{d_{unique}}{\omega_{avg} \cdot \texttt{N}}.
\end{equation}
Notice that low $dv$ values indicate very similar processing times across operation-machine-worker combinations 
which corresponds to many distinct candidate solutions having equal solution quality, e.g., makespan values.



\subsection{Recent work on FJSSP-W}
\label{sec:relatedwork}
The proportion of work on FJSSP-W is still relatively small, to the best of our knowledge. The related problem instances are usually designed for specific real-world applications, individually built on the basis of common FJSSP instances, or randomly generated. While the design process is not always described in a way that can be replicated, the following works introduce benchmarking instances in line with the FJSSP problem description.

An attempt to balance the makespan, maximum worker workload, and total workload of machines using a multi-objective discrete teaching-learning-based optimization approach is performed in~\cite{usman_flexible_2024}. The authors propose 30 FJSSP-W benchmark instances, based on 9 instances from ~\cite{kacem_pareto-optimality_2002} and ~\cite{fattahi_mathematical_2007}. These are extended by differing numbers of fully-flexible workers. The processing times normally distributed around the original processing times.

Another 31 benchmark instances are given in~\cite{gong_non-dominated_2021} relying on FJSSPs from~\cite{kacem_pareto-optimality_2002},~\cite{brandimarte_routing_1993}, and DPpaulli~\cite{dauzere-peres_integrated_1997}.
The FJSSP-W instances are each assigned the same number of workers as there are machines in the original FJSSP, and the processing times of the workers are again normally distributed around the processing times of the machines in the FJSSP instances. The work concentrates on the minimization of the makespan, labor cost, and green production-related factors using a non-linear integer programming model, as well as a non-dominated ensemble fitness ranking algorithm. 

In~\cite{gnanavelbabu_simulation-based_2021}, a simulation-based approach is used to optimize the FJSSP-W and uncertain processing times on 36 self-generated FJSSP-W instances. The processing times of the instances are sampled randomly between the bounds of the already existing processing times of all machines in the underlying FJSSP instances, while the worker options are introduced in accordance with the already present flexibility in the FJSSP instance. The objective considers minimizing both the makespan and the standard deviation of the makespan.

A comparison of a MILP solver and NSGA-II on a distributed FJSSP-W on instances constructed from a subset of the common FJSSP benchmarks (50 instances) and randomly generated instances (8 instances) is conducted in~\cite{luo_distributed_2022}. The initial FJSSP instances from~\cite{hurink_tabu_1994} are extended randomly and divided into three groups of varying problem size. The construction method is not described in greater detail.

In~\cite{usman_job-shop_2024}, the JSSP with worker flexibility is investigated on 40 self-generated instances. The instances are generated randomly in different sizes. The number of workers varies between $2 \leq w \leq 0.6m$ and the processing times for each worker are assigned normally distributed around a mean operation time drawn from the uniform distribution $U(30, 120)$. The work focuses on a multi-objective approach accounting for maximizing the worker workload, minimizing the makespan, and balancing the human energy expenditure.

A dataset of FJSSP-W instances is presented in~\cite{nouri_benchmark_2018} together with a description of the algorithm used for the conversion. It is based on~\cite{fattahi_mathematical_2007}. Across all operations of each job, the available workers are sampled uniformly for each feasible machine assignment. The processing times for each operation-machine-worker combination are sampled uniformly at random (based on the initial operation-machine duration time). While the created instances are available online~\cite{nouri_benchmark_2018}, the details on the determination of the amount of workers, or the upper and lower bounds of the processing times, are not provided.

\subsection{Design of FJSSP-W instances}
\label{sec:design}
This section describes the construction of a rich collection of FJSSP-W instances.
For receiving useful and (partially) realistic FJSSP-W instances, it is supposed that all operations in the production process require the presence of workers. One further assumes that workers perform the same tasks at potentially different speeds and that not all workers can perform every operation with every (or any) of the machines available for the operation.
To obtain the FJSSP-W instances, the given FJSSP instances (cf. Sec.~\ref{sec:benchmarks}) are extended with the necessary information about worker assignments and related processing times.

To build a benchmarking collection of FJSSP-W instances, this paper follows the approach presented by~\cite{nouri_benchmark_2018}. It allows to extend any of the 402 instance from the FJSSP collections in Table~\ref{tab:benchmarksources} with the necessary information to form an FJSSP-W instance.

In the first step, a maximum number $w$ of workers is assigned to each FJSSP instance that needs to be converted. The default value for this is $w=\lfloor 1.5 \cdot m \rfloor$, i.e., the FJSSP-W instance has about $50\%$ more workers than machines. This is justified insofar as most production environments have a larger number of employees than machines in operation.

The second step traverses all jobs in the FJSSP instance sequentially. For each operation-machine combination of a job $J_i$, a random number $w_{i,j,k}$ of eligible workers is sampled from the uniform distribution $U(1,w)$. According to this count, the identifiers of the workers are drawn and the operation-machine-worker combinations receive random processing times $T_{i,j,k,l}$, being distributed around the original machine processing time $T_{i,j,k}$ of the FJSSP,
\begin{equation}
    \forall l \in \mathcal{W}_{i,j,k} : \quad T_{i,j,k,l} \sim   T_{i,j,k} \cdot U\left(lb, ub\right).
\end{equation}
In this context, $lb$ and $ub$ denote the lower and upper limits for deviations in worker processing times on a machine from the initial machine processing times across the operations. The default values are $lb=0.9$ and $ub=1.1$, respectively. The resulting processing times $T_{i,j,k,l}$ are then rounded to the closest integer to comply with the usual specification.  

Using the three parameters $w,lb$, and $ub$, the method extends each FJSSP instance to a randomly generated FJSSP-W instance variant. By varying the parameters and drawing new numbers, any number of FJSSP-W instances can be generated. However, to create the proposed FJSSP-W set, the process is run once for all of the FJSSP instances mentioned in Sec.~\ref{sec:benchmarks}. The 402 random FJSSP-W instances created in this way are fixed in order to ensure a solid benchmark collection for comparable and replicable algorithmic experimentation.  
\begin{figure}[t]
   \centering
   \includegraphics[width=0.95\textwidth]{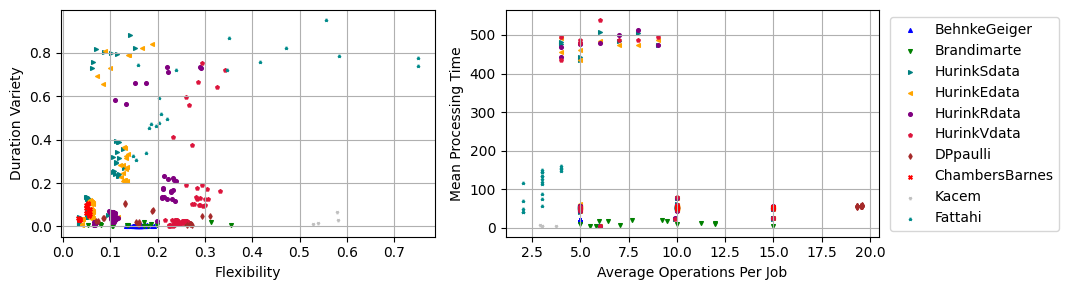}
   \caption{Distribution of selected FJSSP-W instance characteristics. Original FJSSP sources are discriminated using individual colored markers. }
   \label{fig:duration_variation_sources}
\end{figure}

To record the instances, a text file format similar to the well-known FJSSP instances is selected. Details are explained in Section~\ref{sec:fjssp-w_inst}. Analogous to the description in Section~\ref{sec:benchmarks}, the characteristics of the created FJSSP-W instances are documented. This supports the feature-based selection of subsets and the comparison between problem types to obtain improved understanding into the algorithmic working principles of developed solvers.
Figure~\ref{fig:duration_variation_sources} displays the corresponding scatter plots of two pairs of characteristics over all created FJSSP-W benchmark instances. It provides a first idea of the diversity among the instances and reveals where similarities coincide. More details on the individual FJSSP as well as FJSSP-W instances are available in the Github repository of the benchmarking suite~\cite{Hutter_Benchmarking}.

\subsection{FJSSP-W instance description}
\label{sec:fjssp-w_inst}

\begin{figure}[t]
   \centering
   \includegraphics[clip, trim= 0 170 0 0 ,width=0.9\textwidth]{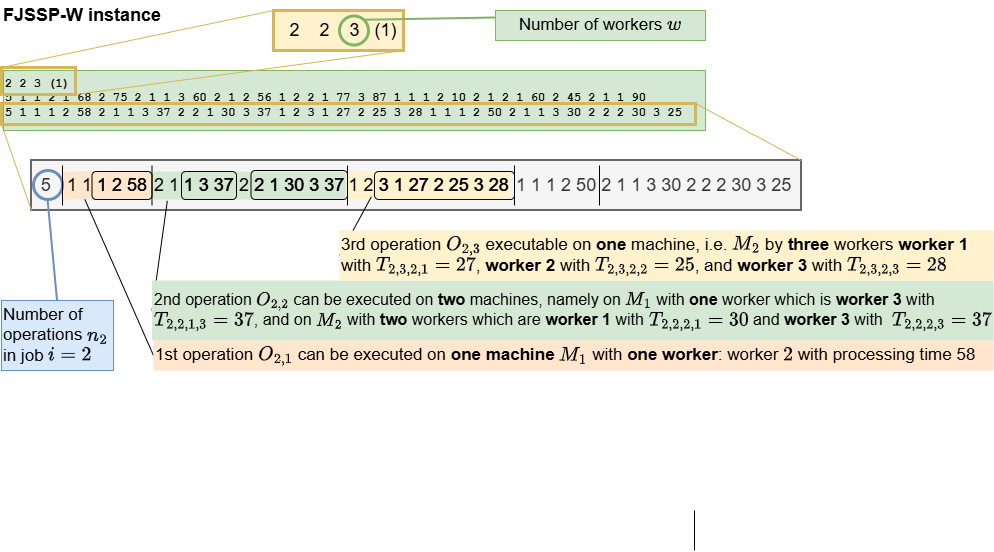 }
   \caption{Explanation of the FJSSP-W benchmark specifications based on an extended FJSSP test instance.}
   \label{fig:textfile_worker}
\end{figure}

The conversion process presented in Sec.~\ref{sec:design} is exemplified in Fig.~\ref{fig:textfile_worker}. It the extends the first row of the original FJSSP instance description with the number of available workers $w=3$ in the third position.
In analogy to the FJSSP format, the following lines represent the individual jobs $J_i$ $(i=1,\dots,n)$ of the FJSSP-W instance to be executed by the production environment.
For each of these jobs, the first value again gives the number of operations $n_i$ to be executed in job $J_i$. 
The second number is the number of machines $m_{i,1}$ available to process the first operation $O_{i,1}$ of job $J_i$.
In the case of the FJSSP-W, each of these machines $\mathcal{M}_{i,1}$  is now equipped with a set of admissible workers $\mathcal{W}_{i,1,k}$. Consequently, the FJSSP instance description format is modified in such a way that the machine-related processing times of the FJSSP instances are replaced by the admissible number of workers for the operation-machine combinations and their corresponding processing times. 
In Fig.~\ref{fig:textfile_worker}, the modifications are highlighted by the black rectangles. The first value inside the rectangles indicates how many different workers can be assigned for the operation-machine combination. This leading number is followed by the corresponding pairs of worker IDs and associated processing times.

Regarding the example in Fig.~\ref{fig:textfile_worker}, the second job $J_2$ in the sample FJSSP-W instance consists of $n_2=5$ operations. The first of these operations, i.e. $O_{2,1}$ can be executed on exactly one of the two available machines. It is machine $M_1$, which can be operated by exactly one of the three workers, namely the second worker $W_2$. Worker $W_2$ can carry out the operation $O_{2,1}$ on machine $M_1$ with processing time $T_{2,1,1,2}=58$.
Analogously, the second operation $O_{2,2}$ of job $J_2$ can be performed on both machines $M_1$ and $M_2$. On machine $M_1$ by one worker, that is, worker $W_3$ with processing time $T_{2,2,1,3}=37$. 
On machine $M_2$ the same operation can be performed by two workers. Worker $W_1$ requires $T_{2,2,2,1}=30$ time units and worker $W_3$ has a processing time of $T_{2,2,2,3}=37$.

\subsection{Modeling of uncertain events}
\label{sec:noise}
{

The benchmark environment also supports simulating uncertainty in the production system. A discrete‑event simulation (DES) models uncertain worker processing times, machine breakdowns, and worker unavailability~\cite{pooch2024discrete,wang2024smart}. Although described for FJSSP‑W, it can be applied to the FJSSP context by assigning each operation a unique fictional worker.

To simulate the uncertainties, a FJSSP-W candidate solution (i.e. a feasible schedule with a certain corresponding makespan) is captured as a directed acyclic graph (DAG). Its vertices represent the operations, while the edges represent the dependencies between the operations. Each vertex can have $0$ to $3$ incoming edges. It may either not depend on prior operations ($0$), or it may possibly have one dependency for the job sequence, for other operations on the same machines, or for other operations assigned to the same worker. If the in-degree is $0$, the vertex is a root vertex (e.g. the first operation of a job which is the first operation on a machine and the first operation that a worker is assigned to in the schedule). 
Equally, each vertex can have $0$ to $3$ outgoing edges. A vertex with out-degree $0$ is considered a leaf vertex (e.g. the last operation of a job, which is also the last operation on a machine and the last of the assigned worker in the given schedule).


\subsubsection{Modeling uncertainty of processing times}


To simulate stochastic processing times, the environment uses multiple probability distributions. In the default case, each worker is assigned their own probability distribution which is used to determine the actual processing time of an operation to emulate inconsistent working speeds encountered in real-life environments. However, the environment also allows for the use of a single distribution and to fix the probability distributions to the machines instead of the workers if necessary to capture a use case. As default, the beta distribution, requiring the shape parameters $\alpha$ and $\beta$, is used to generate the processing time noise~\cite{flores_gomez_monte_2021}.

The noise-free durations $T_{i,j,k,l}$ act as the best-case scenario for the actual durations. The noisy duration $\tilde{T}_{i,j,k,l} = T_{i,j,k,l} (1+\textrm{Beta}(\alpha, \beta))$ is randomly generated in each simulation run with default parameter values $\alpha = U(0,1)$, and $\beta=10\alpha$, respectively. 
These default values cause only minor deviations from the original processing times, e.g., corresponding to smaller human-related inconsistencies.



\subsubsection{Modeling of uncertain resource availabilities}
For the machine breakdowns and the worker unavailability, the environment generates uncertain events and their durations at the start of the simulation run.

\paragraph{Machine Breakdowns}
The uncertain events for machine breakdowns are determined using a exponential distribution. The default probability for the machine breakdowns are designed to produce one machine breakdown per schedule on average, based on the noise-free makespan of the schedule. 
Accordingly, each machine is assigned its own exponential distribution with randomly generated parameter $\lambda_M = \lambda \cdot U(0.9, 1.1)$ centered around $\lambda=\frac{C}{m}$. 
Here, $C$ corresponds to the noise-free makespan of the current candidate solution and $m$ is the number of machines in the related FJSSP instance. 

The duration of each machine breakdown event is determined using a Weibull distribution~\cite{ferrisi_application_2025,xu_multi-stage_2023}. The minimum duration of a machine breakdown by default is set to last at least $10\%$ of the noise-free makespan of the current candidate solution. That  is, the parameters of the  Weibull distribution are by default set to $\alpha = U(\frac{1}{10}C, \frac{1}{5}C)$, and $\beta = 3.602$ respectively. The latter is chosen to ensure that the standard shape closely resembles a normal distribution.

\paragraph{Worker Unavailability}
Similar to the machine breakdowns, the uncertain events concerning the worker unavailability occurrences are also assuming an exponential distribution, and their durations follow a Weibull distribution~\cite{ferrisi_application_2025,xu_multi-stage_2023}. However, while worker unavailability events are tuned to occur with a lower frequency than the machine breakdowns, their duration is set to be significantly longer than machine breakdowns in the default setting of the environment. In the default setting, the minimum duration of a worker unavailability is at $50\%$ of the noise-free makespan, and the probability of their occurrence is tuned to result in $0.5$ events on average for each simulation run.
The default parameters to realize this behavior are analogously build.
As with the machines, each worker has its own distribution, with an modified $\lambda_W = \lambda \cdot U(0.9, 1.1)$ around 
$\lambda = \frac{C}{2 w}$ for the exponential distribution. The durations use the scale parameter $\alpha = U(\frac{1}{2}C, C)$ as well as the shape parameter $\beta = 3.602$ for the Weibull distribution.




\section{Description of the Benchmarking Environment}
\label{sec:benchmark}

 The proposed benchmarking environment is implemented in the Python programming language and offers several APIs. The offered APIs include evaluation functions for the FJSSP and FJSSP-W problem, comparison and visualization of results, and translation between FJSSP and FJSSP-W problem instances. The suite additionally includes well-known benchmark instances as well as their FJSSP-W equivalents, including best-known results for each.
It also offers utility functions to help with loading, selecting, and preparing problem instances.
\begin{figure}[t]\centering
\includegraphics[width=0.99\textwidth]{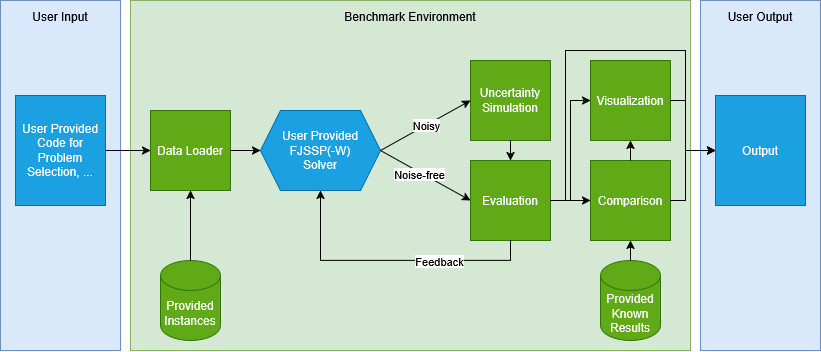}
    \caption{A visualization of the general structure of the benchmarking environment.}
    \label{fig:environment_overview}
\end{figure}
Figure~\ref{fig:environment_overview} illustrates the overall workflow of the developed benchmarking environment.
A detailed description of the implementation and its handling is provided in a public Github repository~\cite{Hutter_Benchmarking}.

\paragraph{Benchmark Selection}
The benchmarking environment offers a filter option that allows benchmark instances to be selected according to the desired benchmark properties. 
The filterable features include the number of operations $\texttt{N}$, the number of machines, the number of workers, the flexibility $\beta$, the duration variety $dv$, the number of orders $n$, and various other features (confer to the documentation~\cite{Hutter_Benchmarking}).

\paragraph{Benchmark Preparation}
Based on the provided filters (if any), the requested benchmarks can be loaded using built-in functions. The loading process returns all requested instances and provides several APIs to read their data.

\paragraph{Benchmarking Experiments}
The benchmarking suite provides best-known results, both for the FJSSP and FJSSP-W instances. These are based on known optima (in case of some FJSSP instances), ro retrieved from the solutions of a collection of state-of-the-art solvers (i.e., Gurobi, IBM ILOG CP Optimizer, or a recent GA variant~\cite{hutter_interior-point_2024}), respectively. All experiments were processed on the same test computer (Intel Core i7-6700 (3.40GHz) CPU with 16 GB RAM, Windows 10) with a time limit of 20 minutes. For the FJSSP-W, only the best-known results for the provided instance set are given. It must be noted that, in case new FJSSP-W instances are generated, these are not anymore comparable to the provided best-known results.

Note that solvers with a stochastic nature like meta-heuristics, multiple algorithm runs per problem instance should be performed to receive conclusive performance results. By default, the minimum recommended number of independent runs is 20. This is a small reduction from the usually recommended number of about 30 repetitions for statistically relevant results~\cite{Vecek2017} and is due to the high computational effort  on the FJSSP(-W) instances.

\paragraph{Performance Evaluation}
The benchmark environment offers an evaluation function for the makespan to ensure the final solution is evaluated correctly. As input for the evaluation, the start times, the machine assignments, and the worker assignments (in case of the FJSSP-W) are required in the fixed order provided by the benchmarking environment. A translation between the encoding used in~\cite{hutter_interior-point_2024} for both the FJSSP and the FJSSP-W is available.
For comparisons to other solvers, the benchmark environment offers different visualizations of the results summarized over all benchmark instances. The data of the evaluated solvers are included and the best known results can be updated over time. Additionally, the benchmarking environment offers an API to calculate the MiniZinc score~\cite{stuckey_minizinc_2014} for comparison.

\paragraph{Uncertainty Simulation}
{In addition to the regular problem instances, the benchmark environment offers the possibility to simulate uncertainties to measure their effects on a given schedule, and to evaluate the effectiveness of uncertainty handling techniques, respectively. The environment uses the discrete event simulation (DES) presented in Sec.~\ref{sec:noise} to introduce uncertainty in processing times, machine breakdowns and worker availabilities within the FJSSP-W, as well as FJSSP, instances. 


The solution is translated to a schedule in form of a DAG accordingly.
To simulate the schedule, the graph is traversed with a modified breadth first search, which prioritizes unexplored parent vertices over child vertices. The traversal starts with all root vertices in the open-list. For each vertex, the processing time of the associated operation is then sampled from the given probability distributions. The new ending time is then determined based on the new starting time, which is the maximum ending time of the parent vertices, or $0$ if the in-degree of the vertex is $0$. This way the new starting and ending times of all operations are guaranteed to be updated in the correct sequence.
}

\subsection{Demonstration of Experiments}
\label{sec:experiments}

The benchmarking environment is demonstrated by comparing different solvers on the FJSSP-W benchmarks.
A complete example (from filtering and loading the data to visualizing the result) comparing the provided Greedy solver with the best known results is provided with the environment. The implementation of the Greedy solver is intended to facilitate the application of the benchmark environments for users. Its results represent absolute baseline results, which should in any case be surpassed on the individual instances.

\subsection{Experimental Settings}

For the FJSSP-W, the solvers were tasked with solving all 402 benchmarks. The hardware used for the experiments is a Intel Core i7-6700 (Quad-Core, 3.40 GHz) CPU with 16 GB RAM. All solvers were given a maximum of 1200 seconds (20 minutes) for each benchmark instance. The respective benchmark instances, including all final results, are available online at \cite{Hutter_Benchmarking}.

\subsection{Applied Solvers}

In order to test the benchmarking environment to compare different solvers for the FJSSP-W, a Genetic Algorithm (GA)~\cite{hutter_interior-point_2024}, a Mixed-Integer Linear Programming solver (MILP), as well as a Constraint Programming solver (CP) are compared to a simple Greedy algorithm as baseline. 
Gurobi~\cite{noauthor_gurobi_nodate} is used for the MILP formulation, and the CP solver provided by the CPLEX library, developed by IBM~\cite{noauthor_constraint_2024} for the CP formulation.
The Greedy algorithm considers the next operation of the sequence of every job as next operation to insert into the schedule. The operation with the smallest machine-worker combination is chosen. If multiple operations offer the same fastest time, the operation is chosen randomly between the best options.

\subsection{Results}

\begin{figure}[t]
\centering
\begin{minipage}{0.4\textwidth}%
    \centering
    (a)\includegraphics[width=0.95\textwidth]{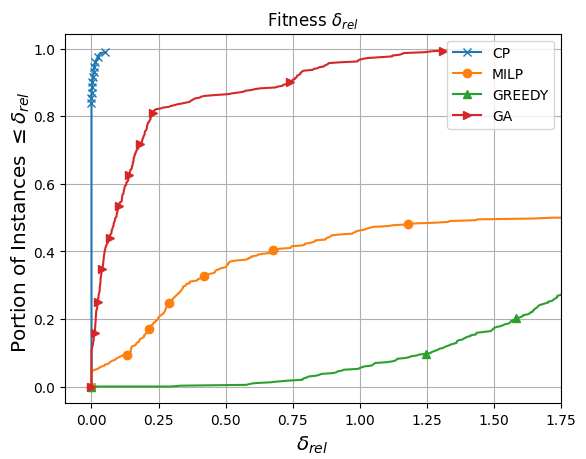}

\end{minipage}%
\qquad
\begin{minipage}{0.5\textwidth}%

\subfloat{
\centering
(b)\includegraphics[clip, trim= 7 26 40 8, width=0.95\textwidth]{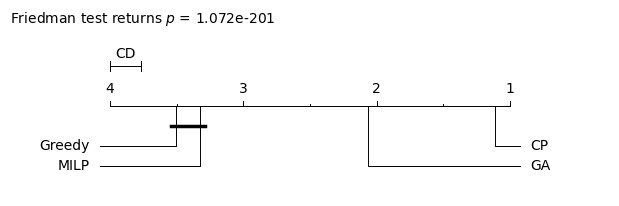}
\label{fig:subfig2}} \\[3ex]

\subfloat{
(c)\quad
        \scalebox{0.8}{
        \begin{tabular}{|r|r|r|r|r|}
        \hline
             \textbf{Solver}  & CP & MILP & GA & Greedy\\ \hline
            \textbf{Score} & 1160.5 & 273.0 & 778.5 & 200.0\\
             \hline
        \end{tabular}
        }
    }
\end{minipage}
\caption{The $\delta_{rel}$ for the FJSSP-W on benchmark instances for the solvers used for the comparison of the approaches is shown in (a). In (b), the Nemenyi diagram for the used solvers is depicted showing the average rankings of each solver. The table in (c) shows the MiniZinc score {
out of a maximum 1206} achieved by the compared solvers.}
\label{fig:greedy_gap}
\end{figure}

\paragraph{Algorithm comparison}
Figure \ref{fig:greedy_gap}(a) shows that the CP as well as the GA approach achieve values of $\delta_{rel} < 0.25$ on more than $80\%$ of the 402 benchmark instances. Yet, neither the MILP solver nor the Greedy approach get anywhere near a similar performance. While the greedy approach would reach $100\%$ eventually, the MILP approach could not solve all problem instances, due to the large amount of variables required and the imposed time limit. Furthermore, hardware limits for large problem instances cause the MILP approach to run out of memory due to the high number of created constraints.
Figure~\ref{fig:greedy_gap}(b) shows that the CP solver significantly outperforms the other approaches. The second placed GA approach also significantly outperforms the MILP and Greedy solvers that do not show significantly different performances over all instances.
The MiniZinc score shown in Fig.~\ref{fig:greedy_gap}(c) also support the results shown in Fig.~\ref{fig:greedy_gap}(b), with the CPs score being considerably higher than the other three solvers, while the GAs score is still significantly higher than the MILP and Greedy score. The MILP and Greedy approach achieve a similar score, with the MILP approach ultimately scoring higher than the Greedy algorithm.

\begin{figure}[t]
\centering
    \subfloat{
    (a)\includegraphics[width=0.45\textwidth]{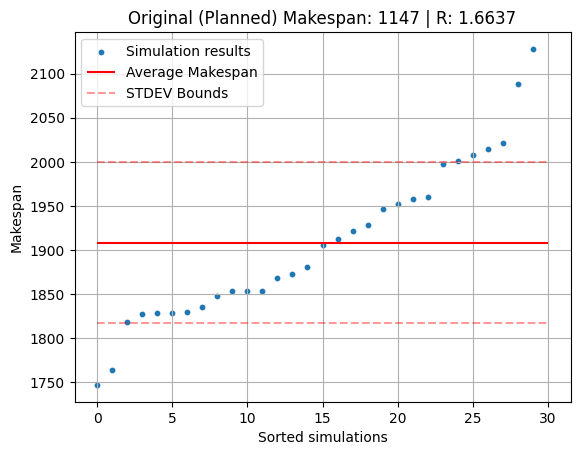}
    (b)\includegraphics[width=0.45\textwidth]{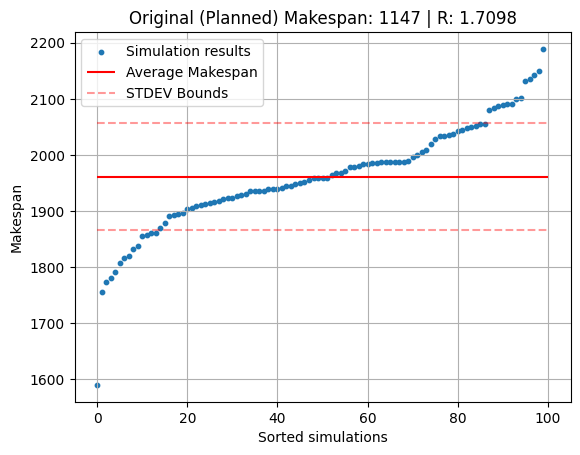}
    }
\caption{A visualization of the simulation results on the example of the problem instance Fattahi20. (a) shows an overview over 30 simulation runs, while (b) shows more accurate results using 100 simulations. Each point in the plots points to the observed real makespan after the simulation accounted for the uncertainties. The red line represents the robust makespan $C_r$ (cf. Eq.~\eqref{eq:robust_makespan}, which represents the expected actual makespan of the provided schedule. The value $R$ in the title indicates the decrease of the makespan from the original plan to the expected makespan (cf. Eq~\eqref{eq:robustness}).}
\label{fig:robust_simulation}
\end{figure}

\paragraph{Robustness}
Using the visualization methods provided by the benchmark environment, Fig.~\ref{fig:robust_simulation} shows an overview of the results of the uncertainty simulation applied to a problem instance. The instance used in this case was the instance Fattahi20. A schedule with the best-known makespan 
 (1147) was used for testing. Fig.~\ref{fig:robust_simulation}(a) shows the results when the schedule is exposed to 30 simulations using the stochastic processing times as only source of uncertainty. Fig.~\ref{fig:robust_simulation}(b) shows more accurate results after 100 simulations. The shown expected actual makespan (displayed as red line) is determined by
\begin{equation}
    C_r = \cfrac{1}{n_{sim}}\sum_{i=1}^{n_{sol}}S_i
    \label{eq:robust_makespan}
\end{equation}
as the average of the results after $n_{sim}$ simulations. The robustness quotient $R$ as shown in the plots title is obtain through
\begin{equation}
    R = \cfrac{C_r}{C},
    \label{eq:robustness}
\end{equation}
where $C$ is the 
noise-free makespan before the simulations account for the used uncertainties. This value can be used to gauge the robustness of the obtained schedule.

\begin{figure}[t]
\centering
    \includegraphics[clip,trim=0 0 0 20, width=0.75\textwidth]{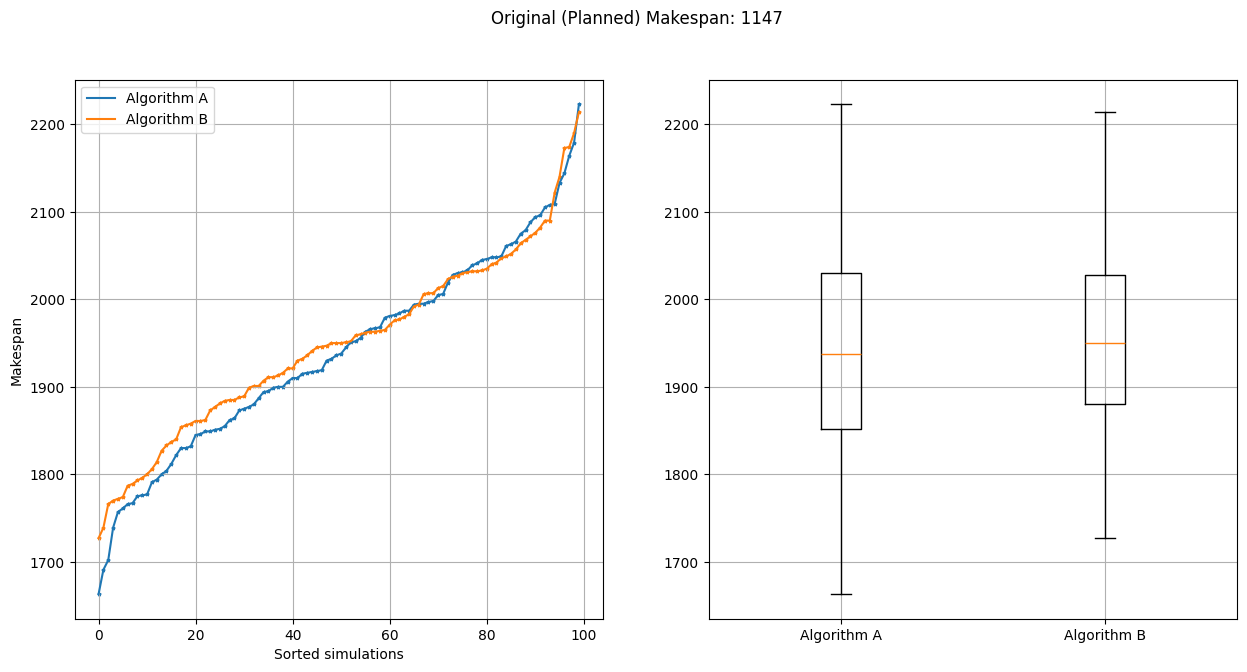}
\caption{An example comparison of two robust results on the same problem instance (Fattahi20). The left plot shows the simulation results for each algorithms obtained schedule, while the right plot shows the distribution of the simulation results. The makespan of the noise-free instance returns a value of 1147.}
\label{fig:robust_comparison}
\end{figure}

An example visualization based on the provided visualization tools in the benchmark environment is shown in Fig.~\ref{fig:robust_comparison}. The comparison can be used to gauge the stability of either schedule to determine their reliability. 

\section{Discussion and Conclusion}
\label{sec:concl}

This paper introduced a comprehensive benchmarking environment for the Flexible Job Shop Scheduling Problem (FJSSP) and its extension, the FJSSP with Worker Flexibility (FJSSP-W). It also offers a method to convert FJSSP instances into FJSSP-W instances. The benchmarking environment enables the comparison of solver results against the best-known solutions. Additionally, it provides tools to filter problem instances based on specific characteristics and to visualize the (relative) performance of various solvers w.r.t. the best-known results of state-of-the-art solvers.

Benchmarking optimization solvers is a critical process that requires a consistent and systematic approach to ensure reliable and insightful algorithm quality assessments. To achieve this, several recommendations~\cite{johnson2002experimental,more2009benchmarking,HELLWIG2019bench} need to be followed and adopted for use of the introduced FJSSP(-W) benchmark environment.

For instance, it is essential to select the most representative set of problem instances that reflect the diversity tied to the benchmarking objectives. To this end, the proposed FJSSP(-W) environment is based on a set of 402 commonly accepted and diverse FJSSP instances. These are systematically extended to account for worker flexibility (FJSSP-W). Although the benchmarking environment includes the code for the independent creation of corresponding FJSSP-W instances, it is strongly recommended to use the fixed and provided 402 FJSSP-W instances of the repository in order to ensure the comparability and replicability of a benchmark suite. This is the only way to generate a comprehensive picture of the performance of a solver with respect to the best known results provided and to future developments.
Further, the creation of additional new instances can be useful in order to investigate certain characteristics in more detail and to expand the space of FJSSP-W test instances in a meaningful way. However, such adaptations should be indicated in the interests of honest scientific work.
The use of the filter to select subsets of the 402 instances is intended in particular for the targeted algorithmic development of operators that focus on specific aspects of the problem. By avoiding the need to evaluate all 402 instances at all times, valuable development time can potentially be saved.

Moreover, choosing appropriate performance metrics is important. To start with, this version of the environment is only designed to measure the makespan. This is a common and useful performance measure for both the FJSSP and the FJSSP-W. In future updates, the environment will be expanded to include further performance measures.
As a computing time limit, we suggest a duration of 20 minutes per algorithm run. This time limit was also used for the present reference results. To grade the computing environment and understand the solver's performance in practical terms, users are expected to report their hardware specifications and the computational resources used.

To ensure that the experiments can be replicated by others, it is necessary to provide detailed descriptions of the setup and parameters. This applies to the parameters of the benchmarking environment (it is recommended to keep the default parameters in Sec.~\ref{sec:benchmark}) as well as for the strategy parameters of the solvers. Whether using default settings or custom configurations, transparency in describing the algorithms and experimental conditions is key. This allows other researchers to reproduce the experiments and verify the results.
The same applies for the uncertainty simulations. As the probability distributions are controlled by the respective parameter choices, either the default parameters should be used or the parameter configuration 
must be used consistently, In any way, the parameter choices must be reported to establish reproducibility and comparability of the performance results.

Since the benchmark environment is aimed at heuristic scheduling procedures and evolutionary algorithms in particular, it is necessary to take the stochasticity of these algorithms into account. This is another important aspect to evaluate the stability of the solver and to understand the reliability and consistency of the results. To this end, developers of such solvers are required to conduct repeated experiments and to analyze the fluctuations in the results. The present paper recommends a minimum amount of $20$ repetitions on each instance to appropriately take into account random effects.
The same stochasticity is present with the uncertainty simulations. These likewise need to be repeated sufficiently often to gather meaningful results. The recommendation for the number of simulations to determine the robust makespan $C_r$ is at least 30, while large problem instances with more resources (worker or machines) could warrant a higher number to ensure reliable results.

The systematic analysis of the results is essential to gain meaningful insights. In order to interpret the results and determine the significance of the differences between the solvers, a statistical analysis should be carried out. In this way, it is possible to determine which solvers perform better under certain conditions, and a solid basis for comparison is created. To ensure this, the present benchmarking environment offers a ranking mechanism that statistically evaluates the performance of several solvers based on the Friedman and Nemenyi test and summarizes it in a corresponding diagram alongside of a visual representation of the overall performance in Fig.~\ref{fig:greedy_gap} to visually support the understanding and communication of the experimental findings effectively. {The environment also provides tools to compare the robustness of multiple solutions to help the understanding of relevant factors present in the different schedules.} Additional visualization tools are available to focus on the analysis of single instances, which are outlined in the provided Github repository~\cite{Hutter_Benchmarking}.

Transparency in reporting benchmarking findings is essential. All relevant code and benchmark data need to be made accessible to the research community. This facilitates subsequent studies and comparisons, contributing to the collective knowledge in the field. Further, the related Github repository ensures continuing support to regularly update the benchmarking processes in consultation with the research community. This helps in refining benchmarking practices and addressing emerging challenges, leading to more accurate and reliable comparisons over time.
Following such guidelines establishes a consistent and systematic approach to benchmarking FJSSP(-W) optimization solvers which will drive progress in this field of research.

\section{Outlook}
\label{sec:outlook}
In future work, additional evaluation metrics (e.g., schedule tardiness) will be added to the benchmarking environment. Additionally, other types of uncertainty to the provided problem instances, like dynamic job arrivals or job cancellations, will be included.

Alongside the closer alignment with real-world use cases, the benchmark environment will address several ways to integrate user interactions, e.g., allowing to assess worst-case effects of certain events, or the fixation of individual jobs in the job sequence and re-optimization of the other jobs.

For easier analysis of the results, additional visualizations might also be added. Considering the different characteristics of the problem instances, an analysis tool indicating which solver performed best on which benchmark characteristics in direct comparison also appears to be a desirable extension.

\section*{Acknowledgements}
The financial support by the Austrian Federal Ministry of Labour and Economy, the National Foundation for Research, Technology and Development and the Christian Doppler Research Association is gratefully acknowledged.

This article is based upon work from COST Action Randomised Optimisation Algorithms Research Network (ROAR-NET), CA22137, supported by COST (European Cooperation in Science and Technology).

 \bibliographystyle{elsarticle-num} 
 \bibliography{cas-refs,mybib,myCEC2023, JournalPaper.bib,newbib,benchmark_suits}







\end{document}